\pdfoutput=1

\documentclass[11pt]{article}

\usepackage{emnlp2021}

\usepackage{times}
\usepackage{latexsym}

\usepackage{graphicx}
\usepackage{amsmath}
\usepackage{dcolumn}
\usepackage{makecell}
\usepackage{array}
\usepackage{enumitem}
\newlength\savedwidth
\newcommand\whline{\noalign{\global\savedwidth\arrayrulewidth
                           \global\arrayrulewidth 1pt}%
                  \hline
                  \noalign{\global\arrayrulewidth\savedwidth}}
\usepackage[T1]{fontenc}

\usepackage[utf8]{inputenc}

\usepackage{microtype}

\title{Bridging the Gap between Language Model and Reading Comprehension: Unsupervised MRC via Self-Supervision}

\author{Ning Bian$^{\rm 1,2}$, Xianpei Han$^{\rm 2}$, Bo Chen$^{\rm 2}$, Hongyu Lin$^{\rm 2}$, Ben He$^{\rm 1,2}$, Le Sun$^{\rm 2}$\\
  $^{\rm 1}$School of Computer Science and Technology, \\University of Chinese Academy of Sciences, Beijing, China \\
  $^{\rm 2}$Institute of Software, Chinese Academy of Sciences, Beijing, China \\
  {\tt \{bianning2019,xianpei,chenbo,linhongyu,sunle\}@iscas.ac.cn} \\ \tt benhe@ucas.ac.cn\\}

\begin{document}
\maketitle
\begin{abstract}
Despite recent success in machine reading comprehension (MRC), learning high-quality MRC models still requires large-scale labeled training data, even using strong pre-trained language models (PLMs). The pre-training tasks for PLMs are not question-answering or MRC-based tasks, making existing PLMs unable to be directly used for unsupervised MRC. Specifically, MRC aims to spot an accurate answer span from the given document, but PLMs focus on token filling in sentences. In this paper, we propose a new framework for unsupervised\footnote{Following \citet{lample2018unsupervised} and \citet{lewis-etal-2019-unsupervised}, we define ``unsupervised'' as ``learning without manually-labeled training data''.} MRC. Firstly, we propose to learn to spot answer spans in documents via self-supervised learning, by designing a self-supervision pretext task for MRC – Spotting-MLM. Solving this task requires capturing deep interactions between sentences in documents. Secondly, we apply a simple sentence rewriting strategy in the inference stage to alleviate the expression mismatch between questions and documents. Experiments show that our method achieves a new state-of-the-art performance for unsupervised MRC.
\end{abstract}

\section{Introduction}

Machine reading comprehension (MRC) aims to answer questions by seeking answer spans in a given document. MRC is widely used to evaluate the language understanding ability of machines \cite{hermann2015teaching,chen-etal-2016-thorough,2016arxiv161101603s,wang2017machine}. Currently, MRC models are still data-hungry. The performances of MRC models depend on large-scale human-generated datasets like SQuAD \cite{rajpurkar-etal-2016-squad}, making it difficult to scale MRC models to new domains and low-resource languages. Therefore, it is critical to develop new data-efficient learning algorithms that do not depend on manually-labeled training data. 

\begin{figure}
   \setlength{\belowcaptionskip}{-0.4cm}
   \centering
   \includegraphics[width=\columnwidth]{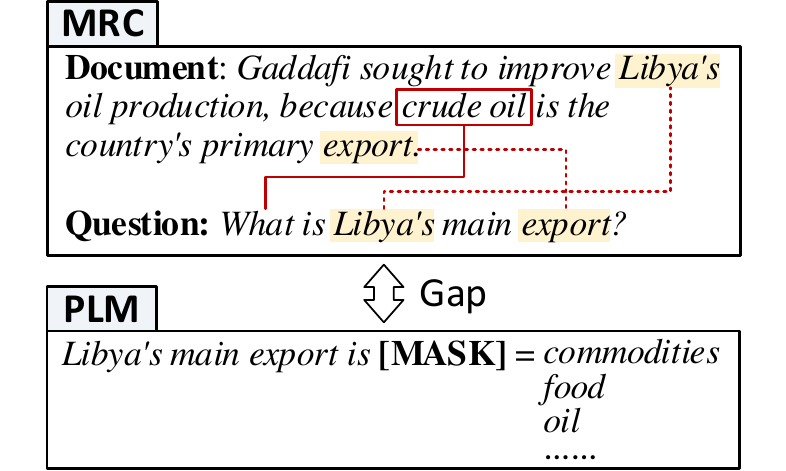}
   \caption{There is a gap between the PLMs and the MRC tasks.}
   \label{figure-f1}
\end{figure}

In recent years, pre-trained language models (PLMs) have achieved great success in many NLP tasks \cite{devlin-etal-2019-bert, radford2019language, 2005.14165}. However, the pre-training tasks for PLMs are not question-answering or MRC-based tasks, so that existing PLMs cannot be directly used for unsupervised MRC. Firstly, the token filling task for PLMs is inconsistent with MRC. As shown in Figure \ref{figure-f1}, MRC requires to spot accurate answer spans in documents with question-document pairs as inputs, while PLMs are usually trained to predict missing/masked tokens in documents. The decoding mechanism for token filling cannot fulfill the requirement of MRC. Secondly, the ability for token filling is different from the ability required by MRC. PLMs prefer to generate suitable words based on the contextual semantics, while MRC requires the context-sensitive span spotting ability to model and exploit deep interactions between question and document. This gap weakens the ability of using existing token-filling-based PLMs for unsupervised MRC.  

\begin{figure*}
   \setlength{\belowcaptionskip}{-0.2cm}
   \centering
   \includegraphics[width=\textwidth]{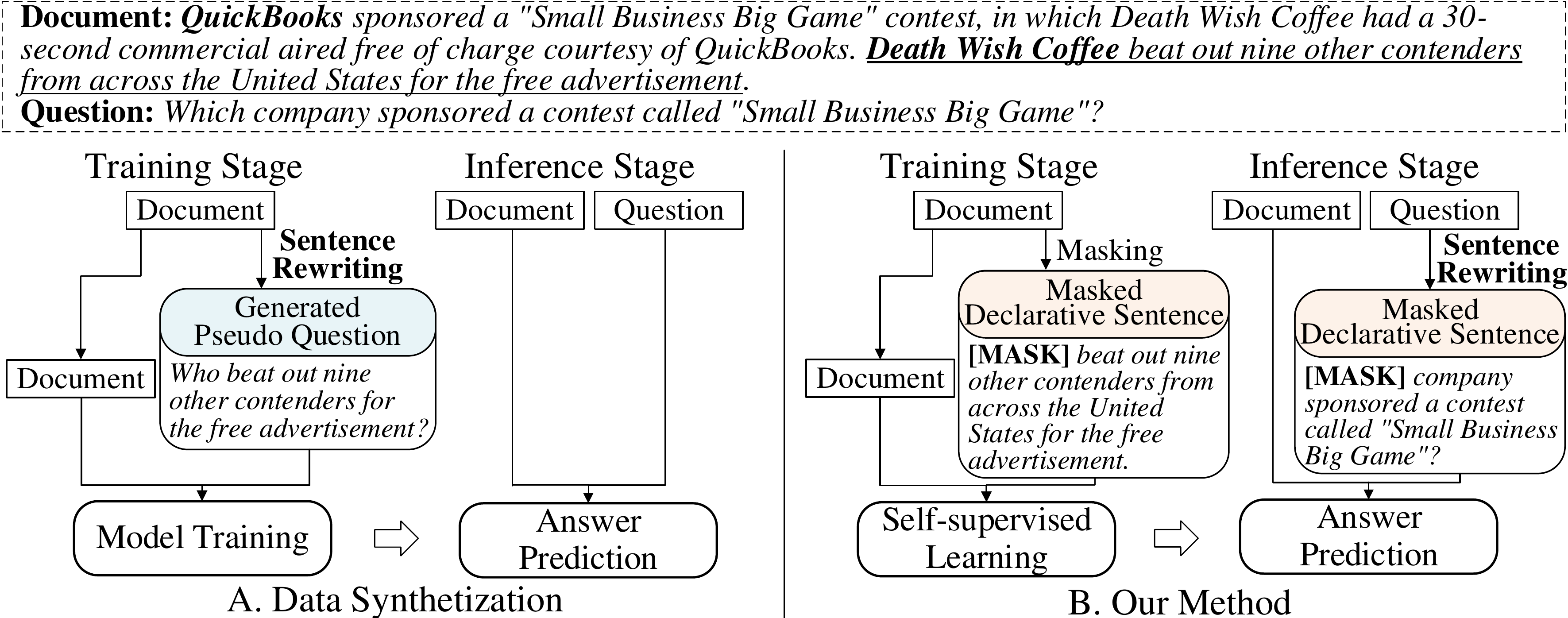}
   \caption{Data synthetization method vs. Our method: Data synthetization methods transform a sentence in the document to question during training. Our method rewrites question into declarative sentence during inference.}
   \label{figure-f2}
\end{figure*}

To fill the above-mentioned gap and adapt PLMs to unsupervised MRC, existing studies leverage data synthetization techniques to generate MRC training data from an unannotated text corpus \cite{lewis-etal-2019-unsupervised,fabbri-etal-2020-template,li-etal-2020-harvesting}. As shown in Figure \ref{figure-f2}A, these data synthetization methods need to rewrite some sentences in documents into pseudo questions during training, because of a mismatch between the expressions of documents in the text corpus (most of the sentences are declarative) and questions in MRC. However, due to the diversity of natural language, neither template-based \cite{fabbri-etal-2020-template} nor data-driven \cite{lewis-etal-2019-unsupervised,li-etal-2020-harvesting} question synthetization methods can generate questions that fully cover the diverse question expressions in real world. This results in an inconsistency between model training and real application situations, limiting the robustness of unsupervised MRC.

In this paper, we propose a new self-supervised learning framework for unsupervised MRC, shown in Figure \ref{figure-f2}B. We propose to learn the context-sensitive span spotting ability via a proper-designed self-supervision pretext task, without any manually-labeled or synthetic training data. In the \textbf{training} stage, we design a new self-supervision pretext task, named Spotting-MLM, to learn the span spotting ability for unsupervised MRC from an unannotated text corpus. Spotting-MLM is a masked language model (MLM) task with a spotting mechanism. Given a passage, we first mask one occurrence of a repeating informative span, and then the MRC model is required to spot and copy a continuous span in the same passage to fill in the mask, which is different from token filling in PLMs. Solving this task relies on capturing deep interactions between the masked sentence and other sentences in the passage. In the \textbf{inference} stage, to alleviate the expression mismatch between questions and documents, we apply a simple sentence rewriting strategy to rewrite the given questions into masked declarative sentences. In this way, our method can rewrite any questions in the real world into declarative forms which are more consistent with the training contexts in self-supervision. This can improve the robustness of our method.

We conduct experiments on four standard MRC datasets in an unsupervised setting. We find that our self-supervised model can effectively learn to spot answer spans in given documents and can significantly outperform data synthetization baselines. Experimental results further show that our framework is complementary to data synthetization methods, as fine-tuning our model on the synthetic datasets can further improve model performance, achieving a new state-of-the-art performance.

The main contributions of our paper are:

1. We propose a new self-supervised learning framework for unsupervised MRC. 

2. Though the Spotting-MLM pretext task, the proposed method can efficiently learn from an unannotated text corpus the span spotting ability for unsupervised MRC.

3. By rewriting questions during the inference stage, the proposed method improves the performance and robustness of unsupervised MRC.
 
\section{Unsupervised Machine Reading Comprehension via Self-supervision}

\begin{figure}
   \setlength{\belowcaptionskip}{-0.4cm}
   \centering
   \includegraphics[width=\columnwidth]{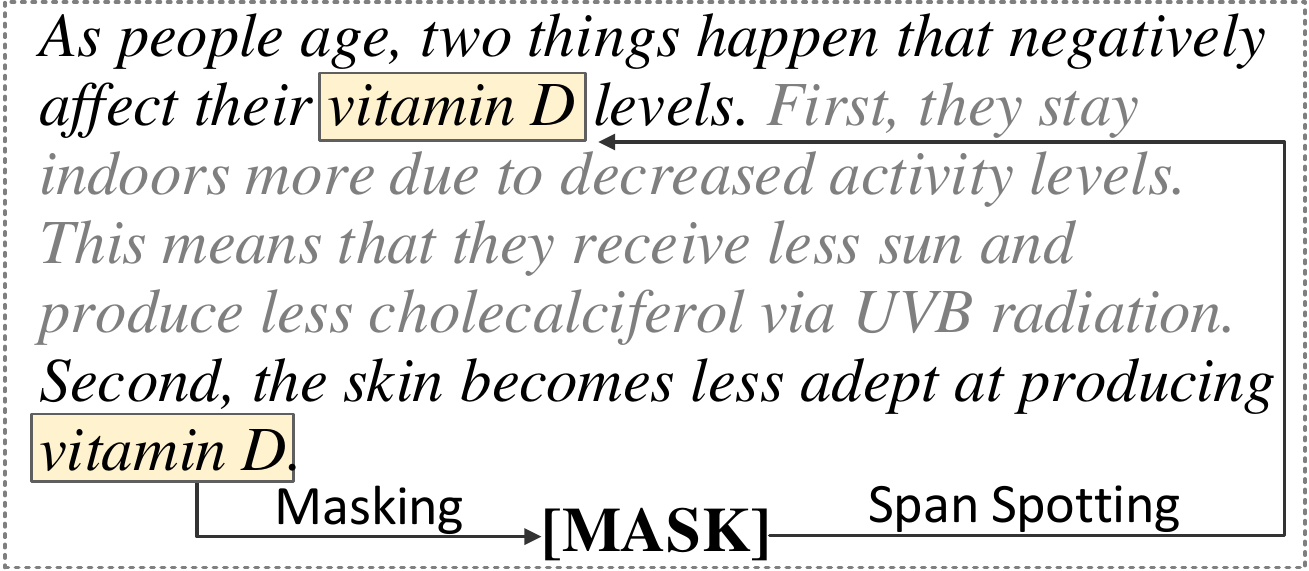}
   \caption{The Spotting-MLM task.}
   \label{figure-f3}
\end{figure}

This section describes our self-supervised learning framework for unsupervised MRC. Given a text corpus, we first construct the self-supervised dataset through informative span masking. Then the dataset is used to train the MRC model through our Spotting-MLM pretext task. During training, we apply a data refinement mechanism for data filtering. In the following, we introduce our self-supervision pretext task, the MRC model, learning details, and the sentence rewriting strategy during inference.

\subsection{Spotting-MLM Pretext Task for MRC}

This section proposes the Spotting-MLM task, which is used to learn unsupervised MRC via self-supervision. As shown in Figure \ref{figure-f3}, given a passage\footnote{In this paper, a passage means ``a paragraph separated by carriage returns''.}, the Spotting-MLM task first masks one occurrence of a repeating informative span (the second ``\textit{vitamin D}'' in this example), and then requires the MRC model to fill in the mask by spotting and copying a continuous span from the passage (i.e., the first ``\textit{vitamin D}''). 

Specifically, the Spotting-MLM task is represented as
\begin{equation}
\setlength{\abovedisplayskip}{0.1cm}
\setlength{\belowdisplayskip}{0.1cm}
\label{equation1}
  p_{mask}\Rightarrow[i_{start},i_{end} ]
\end{equation}
i.e., given a passage with a mask -- $p_{mask}$, the task requires model to predict the start and end positions of a continuous span $[i_{start},i_{end}]$ in the passage to fill in the mask, i.e., $s_{answer}=p_{mask}[i_{start}:i_{end} ]$. For example, in Figure \ref{figure-f3} the second occurrence of ``\textit{vitamin D}'' is masked, and the task requires the model to fill in the mask by spotting the start and end positions of the first ``\textit{vitamin D}'' span as $[i_{start},i_{end}]$. Because the Spotting-MLM task restricts models to fill in the masked part by spotting a continuous span in $p_{mask}$ rather than unconstrainted token prediction, the solving of this task relies on the abilities to exploit deep interactions between the masked sentence and other sentences in the passage and to identify accurate spans, both of which are fundamental to the MRC task. 

Because most humans ask questions to query only informative answer spans, we need to mask meaningful and informative spans in passages. To this end, we propose an informative span masking algorithm. We first find spans that appear \textit{at least twice} in a single passage, like ``\textit{vitamin D}'' in Figure \ref{figure-f3}. Furthermore, we select informative spans as:

1) The spans must appear in the passage at least twice and no more than 4 times. By observation, we find that spans that are too frequent are usually less informative.

2) The spans must contain no stopwords or punctuations, which are not informative in answer spans.

3) The spans must not exceed 5 words in length. If a long span is masked in a sentence, the context of the mask will be less informative.

We discard passages if we cannot find such a span in them. 

After getting the informative span, we mask the last occurrence of the span to match the MRC inference setting, where the question is concatenated after the end of the passage. 

Different from word filling in PLMs that focuses on memorizing correlations between contexts and words, the Spotting-MLM task restricts models to fill in the masked part by spotting and copying an informative span in $p_{mask}$. Through self-supervision, MRC models can learn to spot accurate answer spans in given documents, which relies on capturing interactions between the masked sentence and other sentences in the passage. 

\subsection{MRC Model}
This section describes our model for MRC. Following BERT \cite{devlin-etal-2019-bert}, we model MRC as a mask filling task, i.e., each MRC task is represented as ``[CLS] \textit{Passage} [SEP] \textit{Question} [SEP]'', where the question contains a [MASK] token. 

To capture interactions between the context of the masked token and the answer context, we apply a \textbf{mask-centric span spotting} mechanism: To fill in the mask, the MRC model uses the representation of [MASK] token as the question representation to predict the start and end positions of the answer span in the passage. This is different from the original BERT-based MRC model \cite{devlin-etal-2019-bert}, where the start and end positions of the answer span are predicted by simply applying two fully-connected classifiers upon the representation of each token.

Specifically, the input of our model is a sequence of tokens containing a [MASK] token, denoted as $p_{mask}$. We use BERT \cite{devlin-etal-2019-bert} to get token representations of the sequence:
\begin{equation}
\setlength{\abovedisplayskip}{0.1cm}
\setlength{\belowdisplayskip}{0.1cm}
\label{equation2}
  [\textbf{\textit{x}}_1,\textbf{\textit{x}}_2,...,\textbf{\textit{x}}_i,...]={\rm BERT}(p_{mask})
\end{equation}
where $\textbf{\textit{x}}_i$ is the representation for the $i$-th token, $i$ is the token index.

Then, we use the representation of the [MASK] token, $\textbf{\textit{x}}_{mask}$, as the question representation to predict the start and end positions of the answer span. Firstly, we apply a fully-connected neural network upon $\textbf{\textit{x}}_{mask}$, which is the same as BERT during its MLM pre-training:
\begin{equation}
\setlength{\abovedisplayskip}{0.1cm}
\setlength{\belowdisplayskip}{0.1cm}
\label{equation3}
  \overline{\textbf{\textit{x}}_{mask}}={\rm FCNN}(\textbf{\textit{x}}_{mask})
\end{equation}
Following the MRC model in DrQA \cite{chen-etal-2017-reading}, we use a bilinear term to capture the similarity between each $\textbf{\textit{x}}_i$ and $\overline{\textbf{\textit{x}}_{mask}}$ and compute the probabilities of each token being start and end as:
\begin{equation}
\setlength{\abovedisplayskip}{0.1cm}
\setlength{\belowdisplayskip}{-0.15cm}
\label{equation4}
  {\rm P}_{start}(i|p_{mask})={\rm softmax}(\textbf{\textit{x}}_i{\rm \textbf{W}}_s\overline{\textbf{\textit{x}}_{mask}})
\end{equation}

\begin{equation}
\setlength{\abovedisplayskip}{-0.15cm}
\setlength{\belowdisplayskip}{0.1cm}
\label{equation5}
  {\rm P}_{end}(i|p_{mask})={\rm softmax}(\textbf{\textit{x}}_i{\rm \textbf{W}}_e\overline{\textbf{\textit{x}}_{mask}})
\end{equation}
where ${\rm \textbf{W}}_s$ and ${\rm \textbf{W}}_e$ are learnable parameters. 

The probability of span $[i,i']$ is defined as ${\rm P}_{start}(i|p_{mask})\times {\rm P}_{end}(i'|p_{mask})$. During inference, we choose the span with the highest probability as the final answer.

\subsection{Learning MRC via Self-supervision}
This section describes how to learn the MRC model via the Spotting-MLM task. We first introduce our learning objective, then we design two data refinement strategies: data ranking and iterative data selection.
 
\textbf{Learning Objective}. Given $p_{mask}$ as input, our MRC model is trained to maximize the log probabilities of the correct start and end positions of the answer span. Formally, the learning objective of our self-supervised learning is
\begin{equation}
\setlength{\abovedisplayskip}{0.1cm}
\setlength{\belowdisplayskip}{0.1cm}
\label{equation6}
\begin{split}
  {\rm Spotting\text{-}MLM\,Loss}(p_{mask},i_{start},i_{end})\\
=-\frac{1}{N}\sum_{j=1}^N (log_2{\rm P}_{start}(i_{start}^j|p_{mask}^j)\\
+log_2{\rm P}_{end}(i_{end}^j|p_{mask}^j))
\end{split}
\end{equation}
where $j$ is the index of the training instance, and $N$ is the total number of training instances. 

\textbf{Data Ranking}. To remove the training instances that mask common and trivial spans, we further propose a data ranking strategy based on TF-IDF for data refinement. We calculate three ranking scores for each training instance: 

1) $Score_{Pass}$: To evaluate the informativeness of a passage and prevent meaningless passages, we add up the TF-IDF scores of words in the passage. 

2) $Score_{Mask}$: To evaluate the informativeness of the masked span and prevent adding trivial and meaningless masks, we add up TF-IDF scores of words in the masked span. 

3) $Score_{ans}$: To select better $[i_{start},i_{end} ]$ positions when $s_{answer}$ appears multiple times in $p_{mask}$, we calculate the TF-IDF similarity (ranges from 0.0 to 1.0) between the masked sentence and a window surrounding each occurrence of $s_{answer}$. The window size is set to 10 words.

The final score is a linear combination of the three scores:
\begin{equation}
\setlength{\abovedisplayskip}{0.1cm}
\setlength{\belowdisplayskip}{0.1cm}
\label{equation7}
\begin{split}
  Score =\ &Score_{Pass}\\
+\alpha \cdot &Score_{Mask}\\
+\beta \cdot &Score_{ans}
\end{split}
\end{equation}
where $\alpha$ and $\beta$ are hyperparameters. We rank the self-supervised dataset by the final scores and pick the top-$N$ instances as our training data. Through this ranking, the MRC model can focus on more informative passages and spans that are similar to the real situations in the MRC task.

\textbf{Iterative Data Selection}. In our self-supervised dataset, not all masks are predictable according to their contexts, which is not consistent with the MRC setting and brings biased training signals. To reduce unpredictable training instances, we adopt an iterative data selection strategy, i.e., filtering out instances with low prediction probabilities. Different from the iterative data refinement by \citet{li-etal-2020-harvesting}, we only filter out low-probability training instances and do not refine the answers.

Specifically, we split our training data into 5 groups. Initially, the MRC model is trained on one group, and the trained model is used to calculate the prediction probability of each instance in the next group. We use the probability of answer span $[i_{start},i_{end} ]$ as the prediction probability. Similar to the confidence algorithm by \citet{cao2020unsupervised}, the probability is calculated by softmax over the predicted logits of top-20-scored answer spans. If $[i_{start},i_{end} ]$ is not in the top-20 spans, the probability is set to zero. Then we remove instances whose probability is under a threshold $\tau$. The remaining instances are used for training in the next iteration.

\subsection{Sentence Rewriting during Inference}
There is a mismatch between the expressions of documents in the text corpus (mostly are declarative sentences) and the expressions of the questions in MRC (mostly are interrogative sentences). To alleviate this mismatch, we apply a simple template-based sentence rewriting strategy during the inference stage to rewrite the MRC questions into declarative sentences, as shown in Figure \ref{figure-f2}. 

Specifically, we replace the wh-words in MRC questions with simple templates containing the [MASK] token. For example, ``\textit{Who}'' is replaces by ``\textit{The person} [MASK]'', and ``\textit{Where}'' is replaced by ``\textit{At the place of} [MASK]''. A MRC question ``\textit{Which team won Super Bowl 50?}'' is rewritten to ``[MASK] \textit{team won Super Bowl 50.}''. 

Because we apply sentence rewriting in the inference stage, we only need to generate effective queries for our self-supervised model. Questions after rewriting are more consistent with the training context in self-supervised learning, as the above example shows. Compared with data synthetization methods, we do not need to generate natural language questions from the original documents for MRC training, so that our method doesn't suffer from the question diversity problem arised by sentence rewriting strategies during training.

\section{Experiments}

\subsection{Experiment setup}

\textbf{Datasets}. We conduct experiments on four standard MRC datasets. Because we cannot access the hidden test sets, we use the development sets for evaluation. 

\begin{itemize}[leftmargin=\parindent,itemsep=0pt,topsep=0pt]
\item \textit{SQuAD 1.1} \cite{rajpurkar-etal-2016-squad} is a popular MRC dataset on Wikipedia articles. We use its official dataset.

\item \textit{NewsQA} \cite{trischler-etal-2017-newsqa} contains questions on CNN news articles. We use the dataset provided by \citet{fisch2019mrqa}.

\item \textit{TriviaQA} \cite{JoshiTriviaQA2017} includes question-answer pairs from trivia websites. We use the dataset provided by \citet{sen-saffari-2020-models}, where the passages are Wikipedia articles. 

\item \textit{A Simplified Natural Questions (SimpleNQ)}. NQ \cite{kwiatkowski-etal-2019-natural} uses real questions from Google search. To match the setting of extractive MRC \cite{rajpurkar-etal-2016-squad}, we use the simplified version of NQ provided by \citet{sen-saffari-2020-models}, which is a subset of NQ and the MRC task is defined as finding the short answer in the long answer. 
\end{itemize}

\textbf{Baselines}. We compare our self-supervised method with PLM baselines, Wiki link baselines, and data synthetization baselines.

For PLM baselines, we use PLMs to generate the answer given the concatenation of passage and question as input. The PLM baselines include:

\begin{itemize}[leftmargin=\parindent,itemsep=0pt,topsep=0pt]
\item \textit{BERT} \cite{devlin-etal-2019-bert} is a PLM using MLM task as its pre-training task. We compare with the BERT-Large and BERT-Large with whole-word-masking (wwm) models. 

\item \textit{GPT-2} \cite{radford2019language} is a pre-trained generative language model. 
\end{itemize}

Wiki link baseline: \textit{RefQA} \cite{li-etal-2020-harvesting} leverages citation links in Wikipedia to create document-question-answer triples for MRC training. It is the current state-of-the-art unsupervised MRC model. 

Data synthetization baselines include:

\begin{itemize}[leftmargin=\parindent,itemsep=0pt,topsep=0pt]
\item \textit{\citet{lewis-etal-2019-unsupervised}} generate pseudo questions by unsupervised translation of cloze questions. 

\item \textit{\citet{fabbri-etal-2020-template}} apply template-based question generation on retrieved sentences. 
\end{itemize}

\textbf{Spotting-MLM configuration}. Same as data synthetization baselines, we only use Wikipedia documents as our corpus for self-supervised learning. For data ranking, we set $\alpha=0.5$ and $\beta=200$, and pick top-$N = 500000$ for training. For iterative data selection, we set the threshold $\tau=0.8$. The MRC model is initialized with BERT-Large (wwm) and trained using Adam optimizer \cite{kingma2015adam} with a learning rate of 3e-5 and a batch size of 12. The model is trained for 2 epochs.

\subsection{Main Results}
We conduct unsupervised experiments on the four MRC datasets and measure the performances via the standard Exact Match (EM) and F1 metrics \cite{rajpurkar-etal-2016-squad}. Experimental results are shown in Table \ref{table1} and Table \ref{table2}. We find that:

\renewcommand{\arraystretch}{1.05}
\newcolumntype{d}[1]{D{/}{/}{#1}}
\begin{table*}[t!]
\begin{center}
	\small
    \begin{tabular}{p{10em}|c|d{4}d{4}d{4}d{4}|d{4}}
	\whline
    \textbf{Model}& \multicolumn{1}{c|}{\textbf{External Info.}} & \multicolumn{1}{c}{\textbf{SQuAD}} & \multicolumn{1}{c}{\textbf{NewsQA}} & \multicolumn{1}{c}{\textbf{TriviaQA}} & \multicolumn{1}{c|}{\textbf{SimpleNQ}} & \multicolumn{1}{c}{\textbf{Average}} \\
	\whline
    \textbf{PLM Baselines} & \multicolumn{1}{c|}{}& \multicolumn{1}{c}{} & \multicolumn{1}{c}{} & \multicolumn{1}{c}{} & \multicolumn{1}{c|}{} & \multicolumn{1}{c}{} \\
    \quad GPT-2 & -- -- & 3.1/10.2 & 1.8/8.0 & 5.6/10.0 & 1.7/8.1 & 3.1/9.1 \\
    \quad BERT-Large & -- -- &  5.3/16.8 & 1.4/7.2 & 6.0/14.3 &  3.4/17.9 &  4.0/14.1 \\
    \quad BERT-Large-wwm & -- -- & 12.1/26.2 & 3.1/9.6 & 11.9/22.8 &  7.4/23.1 &  8.6/20.4 \\
	\hline
    \textbf{Wiki link Baselines}& \multicolumn{1}{c|}{} & \multicolumn{1}{c}{} & \multicolumn{1}{c}{} & \multicolumn{1}{c}{} & \multicolumn{1}{c|}{} & \multicolumn{1}{c}{} \\
    \quad RefQA \cite{li-etal-2020-harvesting} &Wiki Doc+Cite link &  57.1/66.8 & 27.4/39.8 & 33.3/41.7 & 39.9/54.5 & 39.4/50.7 \\
    \quad RefQA + Data refine & Wiki Doc+Cite link &  62.5/72.6 & 33.0/46.3 & 40.6/49.4 & 48.7/60.8 & 46.2/57.3 \\
	\hline
    \textbf{Data Synth. Baselines}& \multicolumn{1}{c|}{} & \multicolumn{1}{c}{} & \multicolumn{1}{c}{} & \multicolumn{1}{c}{} & \multicolumn{1}{c|}{} & \multicolumn{1}{c}{} \\
    \quad \citet{lewis-etal-2019-unsupervised} & Wiki Doc &  44.2/54.7^* & - / - & - / - & - / - & - / - \\
    \quad \citet{lewis-etal-2019-unsupervised}\dag &Wiki Doc & 44.3/55.2 & 17.8/27.4 & 19.5/25.4 & 19.2/29.5 & 25.2/34.4 \\
    \quad \citet{fabbri-etal-2020-template} &Wiki Doc &  46.1/56.8^* & - / - & - / - & - / - & - / - \\
    \quad \citet{fabbri-etal-2020-template}\dag &Wiki Doc & 46.2/57.6 & 15.4/24.3 & 27.1/33.7 & 29.8/40.7 & 29.6/39.1 \\
	\whline
    \textbf{Spotting-MLM (Ours)} &Wiki Doc & \textbf{46.4}/\textbf{58.4} & \textbf{18.4}/\textbf{29.6} & \textbf{28.8}/\textbf{37.1} & \textbf{34.5}/\textbf{45.2} & \textbf{32.0}/\textbf{42.6} \\
	\whline
    \end{tabular}%
\end{center}
\caption{\label{table1} Unsupervised results (EM/F1) on MRC datasets and the external information used for MRC learning. ``*'' means results are taken from their original paper. ``\dag'' means our reimplementation. }
\end{table*}%

\renewcommand{\arraystretch}{1.05}
\begin{table*}[t]
  \begin{center}
	\small
    \begin{tabular}{p{18em}|d{4}d{4}d{4}d{4}|d{4}}
    \whline
    \textbf{Model} & \multicolumn{1}{c}{\textbf{SQuAD}} & \multicolumn{1}{c}{\textbf{NewsQA}} & \multicolumn{1}{c}{\textbf{TriviaQA}} & \multicolumn{1}{c|}{\textbf{SimpleNQ}} & \multicolumn{1}{c}{\textbf{Average}} \\
    \whline
    \citet{lewis-etal-2019-unsupervised}  & 44.3/55.2 & 17.8/27.4 & 19.5/25.4 & 19.2/29.5 & 25.2/34.4 \\
    \quad + Spotting-MLM & \textbf{54.2}/\textbf{64.5} & \textbf{22.4}/\textbf{32.7} & \textbf{26.2}/\textbf{33.1} & \textbf{33.3}/\textbf{43.4} & \textbf{34.0}/\textbf{43.4} \\
    \hline
    \citet{fabbri-etal-2020-template}  & 46.2/57.6 & 15.4/24.3 & 27.1/33.7 & 29.8/40.7 & 29.6/39.1 \\
    \quad + Spotting-MLM & \textbf{54.7}/\textbf{65.8} & \textbf{22.2}/\textbf{33.0} & \textbf{29.5}/\textbf{37.1} & \textbf{43.6}/\textbf{55.7} & \textbf{37.5}/\textbf{47.9} \\
    \hline
    RefQA \cite{li-etal-2020-harvesting} &  57.1/66.8 & 27.4/39.8 & 33.3/41.7 & 39.9/54.5 & 39.4/50.7 \\
    \quad + Spotting-MLM & \textbf{59.1}/\textbf{69.7} & \textbf{29.6}/\textbf{42.7} & \textbf{38.1}/\textbf{47.7} & \textbf{40.7}/\textbf{56.8} & \textbf{41.9}/\textbf{54.2} \\
    \hline
    RefQA + Data refine \cite{li-etal-2020-harvesting} &  62.5/72.6 & \textbf{33.0}/\textbf{46.3} & \textbf{40.6}/49.4 & 48.7/60.8 & 46.2/57.3 \\
    \quad + Spotting-MLM & \textbf{64.3}/\textbf{74.9} & 32.9/\textbf{46.3} & 40.4/\textbf{50.0} & \textbf{48.9}/\textbf{62.1} & \textbf{46.6}/\textbf{58.3} \\
    \whline
    \end{tabular}%
\end{center}
\caption{\label{table2} Results (EM/F1) of combining our self-supervised learning method with data synthetization methods.}
\end{table*}%

\renewcommand{\arraystretch}{1.05}
\begin{table*}[t!]
\setlength{\belowcaptionskip}{-0.2cm}
  \begin{center}
	\small
    \begin{tabular}{p{18em}|d{4}d{4}d{4}d{4}|d{4}}
    \whline
    \textbf{Method} & \multicolumn{1}{c}{\textbf{SQuAD}} & \multicolumn{1}{c}{\textbf{NewsQA}} & \multicolumn{1}{c}{\textbf{TriviaQA}} & \multicolumn{1}{c|}{\textbf{SimpleNQ}} & \multicolumn{1}{c}{\textbf{Average}} \\
    \whline
    Full method   & \textbf{46.4}/\textbf{58.4} & \textbf{18.4}/\textbf{29.6} & \textbf{28.8}/\textbf{37.1} & \textbf{34.5}/\textbf{45.2} & \textbf{32.0}/\textbf{42.6} \\
    \whline
    - applying sentence rewriting during training   & 43.6/54.4 & 18.2/28.3 & 27.6/35.3 & 30.6/39.3 & 30.0/39.3 \\
    - w/o informative span masking    & 34.6/47.4 & 12.9/23.0 & 23.2/33.5 & 28.1/43.9 & 24.7/37.0 \\
    - w/o data ranking   & 36.7/48.5 & 14.2/23.5 & 26.0/34.9 & 28.5/42.5 & 26.4/37.4 \\
    - w/o iterative data selection  & 39.6/52.3 & 14.7/26.0 & 25.6/34.9 & 31.0/42.7 & 27.7/39.0 \\
    - w/o mask-centric span spotting  & 45.6/57.7 & 17.9/28.4 & 26.6/35.1 & 33.5/44.5 & 30.9/41.4\\
    \whline
    \end{tabular}%
\end{center}
\caption{\label{table3}Results (EM/F1) of ablation experiments. }
\end{table*}%

\textbf{(1) \textit{PLMs cannot be directly used for unsupervised MRC}.} In Table \ref{table1}, the performances of unsupervised PLM baselines are far from satisfactory: the best baseline BERT-Large-wwm can only achieve 8.6/20.4 EM/F1 scores on average. These results verify that existing PLMs can not be directly used for unsupervised MRC. It is important to design pretext tasks that can precisely train the abilities for down-stream MRC tasks.

\textbf{(2) \textit{Our method is an effective technique for unsupervised MRC}.} 

In Table \ref{table1}, our Spotting-MLM achieves significant improvements compared with PLM baselines. On average F1 score, our model can outperform GPT-2 by 33.5 points, BERT-Large by 28.5 points, and BERT-Large-wwm by 22.2 points. 

As shown in Table \ref{table1}, our method can surpass strong data synthetization baselines on all four MRC datasets. On average F1 score, our model outperforms \citet{lewis-etal-2019-unsupervised} by 8.2 points and outperforms \citet{fabbri-etal-2020-template} by 3.5 points. Moreover, our model obtains larger improvements on the more difficult real-world dataset, SimpleNQ, achieving 15.7 points and 4.5 points improvements, respectively. We believe this is because synthesizing diverse questions that cover the expressions of real-world questions is very challenging, while our method can adapt to different datasets by self-supervisedly learning from a large-scale text corpus and rewriting questions during inference. 

RefQA leverages citation links in Wikipedia in addition to the documents, which provide manually-created, high-quality supervision signals for MRC learning. Consequently, it is unfair to directly compare RefQA with our method. Although our model does not achieve higher performances than RefQA, we show in Table \ref{table2} that our self-supervision method is complementary to RefQA.

\textbf{(3) \textit{Our method is complementary to the data synthetization methods}.} As shown in Table \ref{table2}, using our self-supervised model for initialization and fine-tuning with synthetic datasets can further improve model performances. On average F1 score, we achieve 9.0 points improvement over \citet{lewis-etal-2019-unsupervised}, 8.8 points improvement over \citet{fabbri-etal-2020-template}, and 3.5 points improvement over RefQA \cite{li-etal-2020-harvesting}. We achieve a new unsupervised state-of-the-art result on SQuAD, TriviaQA, and SimpleNQ datasets by combining our self-supervised method and ``RefQA + Data refine'', achieving a 1.0 point improvement on average F1. 

\subsection{Ablation Studies}

In the following, we analyze our method in detail. 

\textbf{Effect of sentence rewriting}. Table \ref{table3} shows the results of applying sentence rewriting during training, which is consistent with the data synthetization paradigm. We use the template-based rewriting strategy by \citet{fabbri-etal-2020-template} to rewrite the masked sentences into questions during training. The performance of using sentence rewriting during training is 3.3 points lower than our method on average F1. This verifies that rewriting sentence during inference is more effective than doing it during training for alleviating expression mismatch between question and document.

\textbf{Effect of informative span masking}. We compare informative span masking strategy with naïve span masking strategy, i.e., all spans appearing at least twice in passages are randomly masked. We can see a severe performance drop in Table \ref{table3}, 5.6 points lower on average F1. This verifies the importance of masking informative spans and demonstrates that our masking strategy is effective.

\renewcommand{\arraystretch}{1.07}
\begin{table}[t!]
\setlength{\belowcaptionskip}{-0.3cm}
  \begin{center}
	\small
    \begin{tabular}{ccccc}
    \whline
    Frequency & 1     & 2     & 3     & $\geq$4 \\
    \hline
    \#Questions & 4184  & 3821  & 1557  & 988 \\
    Proportion & 39.6\% & 36.1\% & 14.9\% & 9.3\% \\
    \hline
    EM    & 53.1  & 43.2  & 36.5  & 46.7 \\
    F1    & 64.3  & 56.1  & 49.2  & 57.2 \\
    \whline
    \end{tabular}%
\end{center}
\caption{\label{table4}Results of our Spotting-MLM model on different answer frequencies of SQuAD dev set. }
\end{table}%

\textbf{Effect of data ranking}. We experiment with randomly sampling the same amount of data without data ranking. As shown in Table \ref{table3}, there is  a significant performance drop compared with the full method (5.2 points drop on average F1). This shows the importance of high-quality self-supervision data and verifies the effectiveness of our data ranking.

\textbf{Effect of iterative data selection}. We remove the iterative data selection and train our model using all data altogether. As shown in Table \ref{table3}, learning without iterative data selection results in a 3.6 points drop on average F1, which verifies the effectiveness of our iterative data selection. 

\textbf{Effect of mask-centric span spotting}. As shown in Table \ref{table3}, the mask-centric span spotting mechanism in our MRC model architecture leads to a 1.2 points performance improvement on average F1, compared with the original BERT-based MRC model \cite{devlin-etal-2019-bert} without this mechanism. This shows the effectiveness of leveraging the mask context information for answer span spotting.

\textbf{Effect of answer frequency}. In Table \ref{table4} we present the performances of our model on different answer span frequencies in the passage. As shown in Table \ref{table4}, there is no positive correlation between the performance and the answer frequency. Compared with cases where answers appear 3-times in the given passages, our model performs better when answers only appear once. This result verifies that our model is not simply identifying high-frequency spans as answers and is robust under different answer frequency conditions. 

\subsection{Adversarial Results}

To further verify the robustness of our self-supervised learning method, we evaluate our model on the adversarial dataset of SQuAD \cite{jia-liang-2017-adversarial}. In this dataset, sentences similar to the questions are appended to the passages to intentionally mislead MRC models to give wrong answers, especially for models that rely on lexical overlapping. We focus on two types of misleading sentences: AddSent and AddOneSent.

\renewcommand{\arraystretch}{1.05}
\begin{table}[!t]
\setlength{\belowcaptionskip}{-0.4cm}
  \begin{center}
	\small
    \begin{tabular}{ld{4}d{4}}
    \whline
    Model & \multicolumn{1}{c}{AddSent} & \multicolumn{1}{c}{AddOneSent} \\
    \hline
    \citet{lewis-etal-2019-unsupervised} & 0.1/9.5 &   1.2/13.6\\
    \citet{fabbri-etal-2020-template} &   2.1/13.9 & 0.4/7.4 \\
    \hline
    Spotting-MLM  & \textbf{30.4}/\textbf{39.9} & \textbf{38.2}/\textbf{48.3} \\
    \whline
    \end{tabular}%
\end{center}
\caption{\label{table5}Results (EM/F1) of baselines and our Spotting-MLM model on the adversarial evaluation set of SQuAD. }
\end{table}%

\renewcommand{\arraystretch}{1.0}
\begin{table*}[!h]
\setlength{\belowcaptionskip}{-0.3cm}
  \begin{center}
	\small
    \begin{tabular}{p{18em}|d{4}d{4}d{4}d{4}|d{4}}
    \whline
    \textbf{Model} & \multicolumn{1}{c}{\textbf{SQuAD}} & \multicolumn{1}{c}{\textbf{NewsQA}} & \multicolumn{1}{c}{\textbf{TriviaQA}} & \multicolumn{1}{c|}{\textbf{SimpleNQ}} & \multicolumn{1}{c}{\textbf{Average}} \\
    \whline
    \textbf{BERT + 1\% Training Data} & 61.9/74.0 & 37.4/54.3 & 46.3/54.3 & 58.9/70.4 & 51.1/63.3 \\
    \hline
    \textbf{Spotting-MLM + 1\% Training Data} & 73.5/83.6 & 41.6/58.4 & 47.6/55.6 & 64.1/75.9 & 56.7/68.4 \\
    \hline
    \textbf{Data Synthetization + 1\% Training Data} &       &       &       &       &  \\
    \quad \citet{lewis-etal-2019-unsupervised} & 68.7/79.5 & 41.5/58.0 & 47.8/55.6 & 63.9/75.6 & 55.5/67.2 \\
    \quad \quad + Spotting-MLM & \textbf{74.3}/\textbf{84.1} & \textbf{43.4}/\textbf{58.2} & \textbf{47.9}/\textbf{55.9} & \textbf{64.3}/\textbf{76.2} & \textbf{57.5}/\textbf{68.6} \\
    \hline
    \quad \citet{fabbri-etal-2020-template} & 68.8/79.9 & 34.8/49.4 & 40.8/48.8 & 63.6/74.9 & 52.0/63.3 \\
    \quad \quad + Spotting-MLM & \textbf{71.3}/\textbf{81.8} & \textbf{35.5}/\textbf{51.0} & \textbf{41.4}/\textbf{49.8} & \textbf{65.5}/\textbf{76.9} & \textbf{53.4}/\textbf{64.9} \\
    \hline
    \quad RefQA \cite{li-etal-2020-harvesting} & 74.6/84.2 & 43.0/59.7 & 48.8/56.5 & 65.0/76.4 & 57.9/69.2 \\
    \quad \quad + Spotting-MLM & \textbf{76.1}/\textbf{85.8} & \textbf{43.4}/\textbf{60.6} & \textbf{48.9}/\textbf{57.0} & \textbf{65.9}/\textbf{77.3} & \textbf{58.6}/\textbf{70.2} \\
    \hline
    \quad RefQA + Data refinement \cite{li-etal-2020-harvesting} & 75.9/85.2 & 43.4/60.3 & 48.2/56.0 & 66.1/77.3 & 58.4/69.7 \\
    \quad \quad + Spotting-MLM & \textbf{76.6}/\textbf{86.1} & \textbf{44.2}/\textbf{61.8} & \textbf{49.4}/\textbf{57.3} & \textbf{67.7}/\textbf{78.9} & \textbf{59.5}/\textbf{71.0} \\
    \whline
    \textbf{BERT + Full Training Data} & 86.8/93.0 & 52.4/68.5 & 60.0/65.5 & 80.5/88.1 & 69.9/78.8 \\
    \whline
    \end{tabular}%
\end{center}
\caption{\label{table6} Low-resource results on MRC datasets using limited human-generated MRC training data. Average performances (EM/F1) over 5 randomly sampled subsets of training data are reported.}
\end{table*}%

As shown in Table \ref{table5}, the data synthetization baselines perform poorly on the adversarial datasets, which demonstrates that these two methods are not robust enough and can be easily misled by noisy passages. 

By contrast, our self-supervised learning method can robustly learn to spot answer spans from documents for MRC. Our self-supervised method achieves a significant improvement compared with data synthetization baselines, achieving 26.0 points and 34.7 points improvements on F1 scores upon the best-performing baselines on these two adversarial datasets.

\subsection{Low-resource Results}

To demonstrate whether our method can be further improved with few training instances, we conduct experiments in a low-resource setting. We use the best configuration of our self-supervised model and then fine-tune it with a limited amount of training data on each MRC dataset. Specifically, for each dataset, we randomly sample 5 subsets of 1\% training data from the training set and report the average performances, shown in Table \ref{table6}.

From Table \ref{table6} we can see that our self-supervised model can outperform the BERT model in the low-resource setting. Moreover, when our self-supervised learning method is combined with data synthetization methods, the low-resource performances can be further improved. The best-performing model can achieve an average F1 of 71.0 using only 1\% of the labeled training data. These results illustrate that our method is effective in further reducing the demand for manually-labeled MRC data.

\section{Related Work}

\textbf{Unsupervised MRC}. Recent unsupervised MRC studies mainly adopt data synthetization approachs \cite{yang2017leveraging,dhingra2018simple,pan2020unsupervised,hong2020handling,kang-etal-2020-regularization}. The main challenge of data synthetization is how to generate real, natural, and diverse questions from documents. \citet{lewis-etal-2019-unsupervised} adopt an unsupervised translation model to transform cloze questions into natural questions. \citet{fabbri-etal-2020-template} retrieve alternative sentences and adopt a template-based question generation method. These two data synthetization methods cannot generate questions that fully cover the diverse question expressions in the real world, resulting in an inconsistency between model training and real applications, which limits the robustness of unsupervised MRC \cite{sultan-etal-2020-importance}. RefQA \cite{li-etal-2020-harvesting} utilizes citation links in Wikipedia to learn unsupervised MRC, which provides manually-created, high-quality supervision signals for MRC learning and limits its generalization to other corpora. 

\textbf{Self-supervised Learning for NLP}. Recent studies have verified the effectiveness of self-supervised learning \cite{raina2007self} for different NLP tasks by designing proper pretext tasks \cite{wang2019self,wu2019self,banerjee2020self,wu2020improving,ruckle2020multicqa,shi2020simple,yamada2020global,xu2020learning,guu2020retrieval}. For extractive summarization, \citet{wang2019self} design pretext tasks including masking, replacing, and switching sentences in passages to learn contextualized representations. On dialogue generation, through the inconsistent order detection task \cite{wu2019self} or utterance prediction and restoration tasks \cite{wu2020improving,xu2020learning}, models can learn to capture relationships between utterances in dialogue flows. On question answering, \citet{banerjee2020self} propose the Knowledge Triplet Learning task to learn multiple-choice QA, and \citet{ruckle2020multicqa} use self-supervision for unsupervised transfer of answer matching ability among domains. \citet{glass-etal-2020-span} propose a span selection task for MRC pre-training. In this paper, we demonstrate that MRC models can be unsupervisedly constructed via self-supervised learning.

\section{Conclusion}

In this paper, we present a new self-supervision framework for unsupervised MRC to bridge the gap between PLMs and MRC. We propose a new self-supervision pretext task, Spotting-MLM, to explicitly learn the context-sensitive span spotting ability for MRC. We apply a sentence rewriting strategy in the inference stage to resolve the expression mismatch between questions and documents. Experiments show that our method is effective in learning unsupervised MRC. Moreover, our method is complementary to the existing data synthesizing methods. Combined with data synthesizing, our method achieves a new state-of-the-art performance on unsupervised MRC.

\bibliographystyle{acl_natbib}
\bibliography{example_output}

\appendix

\renewcommand{\arraystretch}{1.1}
\begin{table*}[!h]
\begin{center}
	\small
    \begin{tabular}{p{11em}p{17em}}
    \whline
    \textbf{Wh-word} & \textbf{Template} \\
    \hline
    A + what + B & A + [MASK] + B \\
    A + how many + B & A + the number of + B + is [MASK] \\
    A + how long + B & A + the length of + B + is [MASK] \\
    A + how much + B & A + [MASK] money + B \\
    A + how + B & A + in [MASK] way + B \\
    A + who + B & A + the person [MASK] + B \\
    A + when + B & A + at the time of [MASK] + B \\
    A + which + B & A + [MASK] + B \\
    A + where + B & A + at the place of [MASK] + B \\
    A + why + B & A + B + because [MASK] \\
    \whline
    \end{tabular}%
\end{center}
\caption{\label{table7}Template for each wh-word. A and B are sentence parts before and after the Wh-words.}
\end{table*}%

\renewcommand{\arraystretch}{1.1}
\begin{table*}[!h]
\begin{center}
    \small
    \begin{tabular}{p{16.61em}cccc}
    \whline
    \multicolumn{1}{l}{} & \textbf{SQuAD} & \textbf{NewsQA} & \textbf{TriviaQA} & \textbf{SimpleNQ} \\
    \hline
    Number of development data & 10550 & 4212  & 9835  & 2356 \\
    Number of training data & 87599 & 92549 & 77084 & 74218 \\
    Average number of tokens in questions & 12.3  & 8.2   & 19.1  & 9.8 \\
    Average number of tokens in documents & 155.8 & 825.4 & 844.6 & 126.3 \\
    Average number of tokens in answers & 4.2   & 5.9   & 2.2   & 5.8 \\
    \whline
    \end{tabular}%
\end{center}
\caption{\label{table8}Details of the MRC datasets used in our experiments.}
\end{table*}%

\section{Details of Sentence Rewriting}

In Table \ref{table7}, we provide the template for each wh-word. For questions that cannot find wh-words, we simply add a [MASK] at the end of the questions.

\section{Details of MRC Datasets}
Details of the MRC datasets used in our experiments are shown in Table \ref{table8}. Note that ``tokens" in Table \ref{table8} are wordpieces in BERT \cite{devlin-etal-2019-bert}.

\section{Reproducibility Information}
For training our MRC model, we randomly sample a subset from the synthetic training datasets \cite{lewis-etal-2019-unsupervised,fabbri-etal-2020-template} as our development set for hyperparameter tuning and model selection.A linear decay of learning rate is applied during training. The max sequence length is set to 384, and we split long documents into multiple windows with a stride of 128. 

We use the official evaluation code of SQuAD \cite{rajpurkar-etal-2016-squad} for our evaluations, available at ``https://rajpurkar.github.io/SQuAD-explorer/''.

We conduct all our experiments on a Nvidia TITAN RTX GPU using TensorFlow 1.14.0.
 
\section{Baseline Implementation Details}
For the BERT baselines, we first rewrite questions to masked sentences and use BERT to generate the masked parts as answers. For answers with multiple words, we adopt an iterative generation approach, following \citet{schick-and-schutze2020}. Because we do not know the lengths of the answers, we generate answers with different lengths (maximum length is 5 words) and choose the answer with the highest generative probability as the final answer.

For \citet{lewis-etal-2019-unsupervised} and \citet{fabbri-etal-2020-template} models, we reimplement their model by training on their synthetic data on GitHub. For \citet{li-etal-2020-harvesting}, we use their released model on GitHub for experiments.

\end{document}